\documentclass[10pt,twocolumn,letterpaper]{article}

\usepackage[pagenumbers]{iccv} % To force page numbers, e.g. for an arXiv version
\usepackage{multirow}
\usepackage{booktabs} % For professional tables
\usepackage{pifont}    % For \xmark (cross symbol)

\usepackage{xcolor}    % For colored text
\usepackage{bbm}

\newcommand{\cmark}{\ding{51}}  % Define checkmark symbol
\newcommand{\xmark}{\ding{55}}  % Define cross symbol
\definecolor{iccvblue}{rgb}{0.21,0.49,0.74}
\usepackage[pagebackref,breaklinks,colorlinks,allcolors=iccvblue]{hyperref}

\def\paperID{14486} % *** Enter the Paper ID here
\def\confName{ICCV}
\def\confYear{2025}

\title{Towards Robust Multimodal Representation: A Unified Approach with Adaptive Experts and Alignment}

\author{Nazanin Moradinasab\\
School of Data Science\\
University of Virginia\\
{\tt\small nm4wu@virginia.edu}
\and
Saurav Sengupta\\
School of Data Science\\
University of Virginia\\
{\tt\small ss4yd@virginia.edu}
\and
Jiebei Liu\\
Systems and Information Engineering\\
University of Virginia\\
{\tt\small mcu2xn@virginia.edu}
\and
Sana Syed\\
Division of Pediatric Gastroenterology\\
Duke University Medical Center\\
Duke Clinical Research Institute\\
{\tt\small sana.syed@duke.edu}
\and
Donald E. Brown\\
School of Data Science\\
University of Virginia\\
{\tt\small deb@virginia.edu}
}

\begin{document}
\maketitle
\begin{abstract}
Healthcare relies on multiple types of data, such as medical images, genetic information, and clinical records, to improve diagnosis and treatment. However, missing data is a common challenge due to privacy restrictions, cost, and technical issues, making many existing multi-modal models unreliable. To address this, we propose a new multi-model model called Mixture of Experts, Symmetric Aligning, and Reconstruction (MoSARe), a deep learning framework that handles incomplete multimodal data while maintaining high accuracy. MoSARe integrates expert selection, cross-modal attention, and contrastive learning to improve feature representation and decision-making. Our results show that MoSARe outperforms existing models in situations when the data is complete. Furthermore, it provides reliable predictions even when some data are missing. This makes it especially useful in real-world healthcare settings, including resource-limited environments. 
% Our code is publicly available at \url{https://github.com/NazaninMn/MoSARe}. 
Our code is publicly available at \href{https://github.com/NazaninMn/MoSARe}{MoSARe}. 
\end{abstract}
\vspace{-0.5cm}
\section{Introduction}
\label{sec:intro}

Multimodal data plays a crucial role in healthcare by integrating information from different sources such as medical images, genomic data, clinical notes, and physiological signals to provide a more comprehensive understanding of the patient's health \cite{zhang2022m3care}. Physicians often rely on these different modalities to make well-informed decisions about diagnosis, and treatment. For example, cancer subtyping requires combining histopathology images, genetic profiles, and clinical records to accurately classify cancer types and determine the most effective treatment strategies \cite{wu2024multimodal}.

However, a major challenge in multimodal learning is missing data. In real-world healthcare settings, patient records are often incomplete due to privacy restrictions, financial costs, variations in hospital protocols, or technical issues \cite{wu2024deep}. These challenges are even more pronounced in rural or under-resourced areas, where access to advanced medical technologies may be limited. Many existing multimodal models assume that all data is available, which makes them unreliable when faced with missing information \cite{xue2022modality, ding2024multimodal}. Traditional solutions, such as data imputation \cite{shang2017vigan}, often depend on strong assumptions about data distribution and fail to fully capture inter-patient relationships, which can limit their effectiveness \cite{wu2024multimodal}.

To address this issue, we developed a unified framework for multimodal learning, called Mixture of Experts, Symmetric Aligning, and Reconstruction (MoSARe), designed to handle incomplete multimodal data while maintaining high diagnostic accuracy. Specifically, our approach focuses on designing a multimodal deep learning model for cancer subtyping, integrating biopsy data, RNA-seq data, and text reports, even when certain modalities are missing.

While developing MoSARe, we incorporated cross-modal attention mechanisms and decoupled reconstruction strategies inspired by MECOM \cite{wei2024mecom} to enhance multimodal feature integration and missing data handling. However, directly applying existing multimodal learning frameworks to healthcare presented significant challenges, particularly in processing diverse medical modalities and improving feature representation for complex clinical data.
One of the key challenges was modality-specific preprocessing, especially for medical images. Unlike general image datasets, histopathology images and other complex medical data require domain-specific feature extraction techniques. MECOM relied on CNN-based feature extraction to obtain global and local representations, but this approach proved insufficient for medical imaging, which demands a more structured and clinically meaningful representation. To overcome this limitation, we introduced a specialized preprocessing pipeline, tailored for healthcare modalities, to ensure effective feature extraction and improve downstream learning.
Beyond preprocessing, we further refined and enhanced multimodal learning by integrating adaptive expert selection, symmetric aligning, and multi-prototype contrastive learning, all of which contribute to better feature representation and improved performance in healthcare applications. Mixture of Experts (MoE) enables adaptive feature selection, ensuring that the most informative local representations are utilized for robust learning, even when some modalities are missing. Symmetric aligning improves modality consistency, reducing discrepancies between feature representations extracted from different medical data sources. Additionally, multi-prototype contrastive learning strengthens feature discriminability, refining representations for more precise classification and better generalization across diverse and complex healthcare datasets.
Our proposed model outperforms existing approaches by achieving higher diagnostic accuracy and robustness in multimodal learning when all modalities are available, while also excelling in real-world scenarios where data is often incomplete. By effectively handling missing modalities, MoSARe ensures equitable access to high-quality AI-driven diagnostics, making it particularly valuable in resource-limited healthcare settings. Its ability to align heterogeneous features and learn richer representations enhances adaptability to diverse medical datasets, ensuring reliable and fair clinical decision-making across different patient populations. Therefore, our main contributions are:
% We summarise our main contributions as: 
\begin{itemize}
    \item A Mixture of Experts (MoE) mechanism that enables adaptive selection of the most informative local representations as well as multimodal representation integration.
    % at the end, improving the performance of the model.
    \item Effective handling of missing modalities in multimodal learning without relying on strong assumptions.
    \item Integrating three different modalities, including histopathology images, RNA-seq data, and clinical text.  
    % enhancing decision-making in healthcare applications.
    \item Integrating alignment approaches including symmetric aligning and multi-prototype contrastive learning to improve model classification accuracy.
    \item The proposed approach surpasses state-of-the-art models in complete data scenarios.
\end{itemize}
% 1) Our model introduces a Mixture of Experts (MoE) mechanism that enables adaptive selection of the most informative local representations as well as multimodal representation integration at the end, improving the performance of the model, 2) The proposed model effectively handles missing modalities in multimodal learning without relying on strong assumptions, 3) The framework integrates three different modalities, including histopathology images, RNA-seq data, and clinical text, enhancing decision-making in healthcare applications, 4) Our model integrates the aligning approaches including symmetric aligning and multi-prototype contrastive learning to improve model classification accuracy, 5) The proposed approach surpasses state-of-the-art models in complete data scenarios.

\section{Related Works}
\label{sec:relatedworks}
Recent advances in multimodal learning have demonstrated significant potential in integrating histology and genomic data for comprehensive cancer analysis. Early approaches in this domain focused on late fusion strategies, such as vector concatenation\cite{mobadersany2018GSCNN} and Kronecker product\cite{chen2022pathomicbp}, to fuse histology features with molecular profiles. While these methods demonstrated initial success, they lacked mechanisms to model fine-grained cross-modal interactions and address missing modalities—a pervasive issue in real-world clinical settings. Subsequent works adopted early fusion frameworks, such as transformer-based architectures like MCAT \cite{chen2021multimodal} and MGCT \cite{liu2022mgct}, which utilized co-attention mechanisms to model histology-genomic correlations. However, these frameworks often assume full modality availability, limiting their robustness in scenarios with missing data. 
To address missing modalities, traditional imputation methods\cite{shang2017vigan} often introduced biases and failed to leverage inter-patient relationships, while graph-based methods\cite{zhang2022m3care, you2020handling} attempted to model patient-modality interactions but struggled with combinatorial missing patterns and computational complexity. Recent approaches like MECOM \cite{wei2024mecom} proposed reconstruction-based techniques to handle incomplete multimodal data, yet their reliance on generic CNN-based feature extraction proved inadequate for medical imaging, which demands domain-specific representations. Meanwhile, Mixture-of-Experts (MoE) architectures have shown promise in adaptive feature selection, dynamically routing inputs through specialized subnetworks based on modality availability \cite{zhang2023moe}. However, existing MoE frameworks in healthcare often oversimplify cross-modal alignment or neglect clinically meaningful representations.  These limitations underscore the need for a unified framework that integrates MoE-driven adaptability, symmetric cross-modal alignment, and contrastive representation learning—capabilities critical for robust cancer subtyping in both complete and incomplete data scenarios.
\section{Method}
\label{sec:method}

In this section, we introduce the key components for our proposed framework for robust multimodal learning. The model consists of five main modules: Modality Pre-Processing, Adaptive Cross-Modal Attention, Decoupled Reconstruction for Handling Missing Data, Feature Aligning (to harmonize representations across modalities and refine feature embeddings for more precise classification), and Final Feature Aggregation.
\begin{figure*}[t]
  \centering
    \centering
    \includegraphics[width=0.78\linewidth]{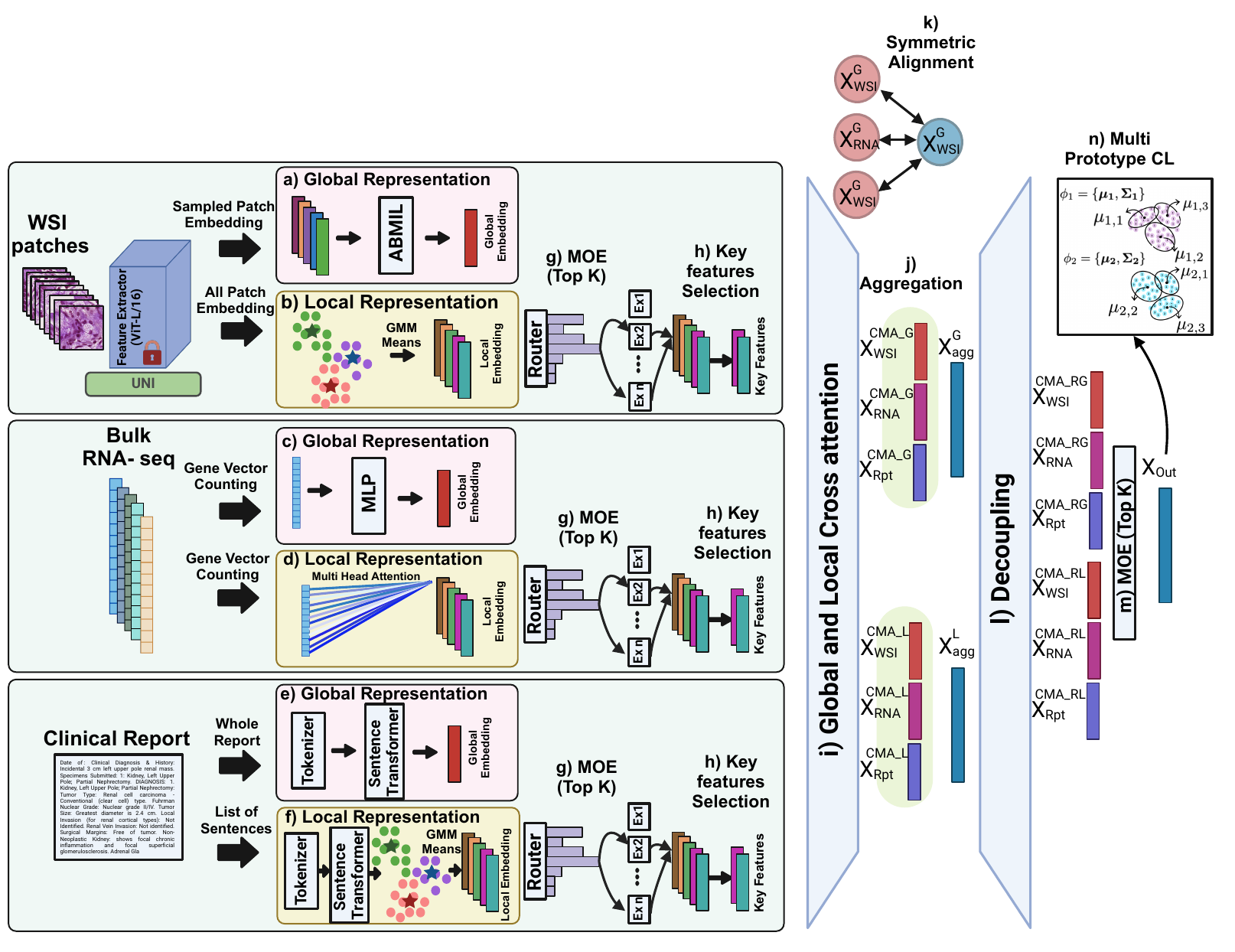} % Ensure the file path is correct
    \caption{The schematic representation of MoSARe}
    \label{fig:model}
\end{figure*}
\subsection{Modality Pre-Processing}

To ensure effective multimodal representation learning, we process and extract global and local feature embeddings from different healthcare data sources, such as histopathology images, RNA-seq data, and clinical reports.

\textbf{Histopathology images:}
A whole slide image (WSI) is divided into \( N_h \) non-overlapping patches of size \( 256 \times 256 \) pixels at 20× magnification (0.5 µm/pixel), forming a set of histology patches \( \{x_{i,h}\}_{i=1}^{N_h} \). Each patch \( x_{i,h} \) is mapped to a low-dimensional feature representation using a pretrained patch encoder \( f_{\text{enc}}(\cdot) \), shown in Eq. \ref{Eq1}:
\begin{equation}
z_{i,h} = f_{\text{enc}}(x_{i,h}) \in \mathbb{R}^D
\label{Eq1}
\end{equation}
We used UNI \cite{chen2024towards}, a ViT-L/16 DINOv2 model \cite{dosovitskiy2020image, oquab2023dinov2} with 1024-dimensional embeddings, pre-trained on a large histology dataset.
In our MoSARe model, we extract both global and local representations to capture high-level contextual patterns and fine-grained local variations in WSIs. As illustrated in Fig. \ref{fig:model}a and \ref{fig:model}b, we employ two distinct strategies to obtain these representations. For global representation (Fig. \ref{fig:model}a), we aggregate patch embeddings into a global representation using attention-based multiple instance learning (ABMIL) \cite{ilse2018attention}. 
Given a set of patch embeddings \( \{z_{i,h}\}_{i=1}^{N_h} \) extracted from a WSI \( h \), the global representation is defined as \( X_{\text{WSI}}^{G} = f_{\text{agg}}(\{z_{i,h}\}_{i=1}^{N_h}) \in \mathbb{R}^D \), where \( f_{\text{agg}}(\cdot) \) is the aggregation function learned by the ABMIL model. This function computes a weighted combination of patch embeddings, ensuring that the most informative tissue regions contribute more significantly to the final global representation.
For local representation (Fig. \ref{fig:model}b), we employ a Gaussian Mixture Model (GMM)-based approach. We first apply K-means clustering to the feature embeddings of all patches across WSIs to obtain cluster centroids, which are then used to initialize a GMM per WSI. The means of the Gaussian components for WSI \( h \) serve as local representations, defined as \( X_{WSI}^{L} = \{x_{j} \mid j = 1, \dots, C\} \), where \( x_{j} \) represents the mean of the \( j \)-th Gaussian component and $C=16$. These components capture tissue diversity, cellular heterogeneity, and spatial organization within each WSI, effectively summarizing local histological variations.
% By integrating global (\( X_{WSI}^{G} \)) and local (\( X_{WSI}^{L} \)) representations, our approach efficiently summarizes the large variable-length set of patch embeddings into a structured feature space, enabling the model to learn both broad contextual features and fine-grained morphological variations, which are crucial for accurate and robust healthcare analysis.

\textbf{RNA-Seq data:}
Following \cite{jaume2024transcriptomics, jaume2024modeling}, we process RNA sequencing data by computing the log\textsubscript{2} fold change relative to a control group to capture gene expression deviations. Given a sample \( i \), the normalized transcriptomic data is represented as \( t_i \in \mathbb{R}^{N_G} \), where \( t_i \) denotes the gene expression vector for sample \( i \), and \( N_G \) represents the total number of genes. To contextualize these gene-level measurements within biological processes, we integrated pathway-level information from Reactome\cite{gillespie2022reactome} and MSigDB-Hallmarks\cite{Liberzon2015msigdb, Subramanian2005genome}curating 331 pathways (281 from Reactome, 50 from Hallmarks) encompassing 4,999 genes after filtering for $\geq 90\%$ transcriptomic coverage.

In our framework, we extract both global and local representations for RNA-seq data, similar to our approach for WSIs, as illustrated in Fig. \(\ref{fig:model}\)c and \(\ref{fig:model}\)d. The global representation is obtained using a three-layer multilayer perceptron (MLP), while the local representation is extracted through an attention-based multiple instance learning (ABMIL) model with $n_{h}$ attention heads which is set to 16.
To obtain the global embedding, we transform RNA-seq feature embeddings into a 1024-dimensional (D) representation using a function $\phi(\cdot): \mathbb{R}^{N_G} \to \mathbb{R}^{D}$, which is implemented as a three-layer MLP (i.e. $X_{RNA}^{G} = \phi(t_i)$). where \( X_{RNA}^{G} \in \mathbb{R}^{D} \) represents the global transcriptomic embedding of sample \( i \). Each layer of the MLP consists of a fully connected layer followed by LayerNorm, ReLU activation, and Dropout (0.2 probability) to prevent overfitting. 
% This hierarchical feature transformation allows the model to effectively capture transcriptomic patterns while reducing noise, providing a robust molecular representation for downstream analysis.

To extract local representations, we employ a multi-head attention-based ABMIL model that learns patch-level attention weights across RNA-seq feature tokens. The model assigns importance scores to different gene subsets and aggregates them into local representations. 
The local representations are defined as \( X_{\text{RNA}}^{L} = \{ x_{j} \mid j = 1, \dots, N_{h} \}, \quad x_{j} = \psi_j(t) \), where \( N_{h} \) represents the number of attention heads, and \( X_{\text{RNA}}^{L} \) denotes the set of \( N_{h} \) local embeddings. \( N_{h}\) is set to 16. 
Each embedding corresponds to a distinct RNA-seq subspace learned through the multi-head attention mechanism, effectively capturing fine-grained transcriptomic variations.
The transformation function \( \psi_j(\cdot) \) for attention head \( j \) is defined in Eq. \ref{Eq:1}.
\begin{equation}
\psi_j(t) = \sum_{k=1}^{N_G} \alpha_{j,k} W_j t_{k}
\label{Eq:1}
\end{equation}
where \( W_j \) is the learned projection matrix for head \( j \), and \( \alpha_{j,k} \) represents the learned attention weight for gene \( k \) in head \( j \). 
% These localized representations help preserve fine-grained transcriptomic variations, making them particularly useful for capturing subtle molecular differences across samples.
% By integrating both global (\( X_{\text{RNA}}^{G} \)) and local (\( X_{\text{RNA}}^{L} \)) representations, our approach ensures a comprehensive characterization of RNA-seq data. This enables the model to learn broad transcriptomic patterns while retaining localized feature variations, which are crucial for accurate and robust biomedical analysis.

\textbf{Clinical reports:}
We first removed all mention of the cancer subtype in the report using string search and replacement, to avoid label leakage. To obtain both global and local representations, as illustrated in Fig.~\ref{fig:model}e and \ref{fig:model}f, we first fine-tune a sentence transformer using ModernBERT \cite{modernbert} to encode textual data at different levels of granularity. For the global representation, we encode the entire clinical report as a single embedding using the sentence transformer \( s_{\text{enc}}(\cdot) \), mapping the full text into a dense vector representation such that \( X_{RPT}^{G} = s_{\text{enc}}(R) \), where \( X_{RPT}^{G} \in \mathbb{R}^{D} \) represents the global embedding. For the local representation, we first apply the sentence transformer at the sentence level to obtain individual sentence embeddings. Given a clinical report \( R \) consisting of \( N_s \) sentences \( \{ S_{j} \}_{j=1}^{N_s} \), each sentence \( S_{j} \) is mapped to an intermediate representation \( s_{j} = s_{\text{enc}}(S_{j}) \), where \( s_{j} \in \mathbb{R}^{D} \). After obtaining sentence embeddings, we follow the same WSI's local representation extraction process to get text local representations using GMM. The local representations are denoted as \( X_{RPT}^{L} \), where $X_{RPT}^{L} = \{ x_{j} \mid j = 1, \dots, C \}, \quad x_{j} \in \mathbb{R}^{D}$. We set $C$ to 16.
% Specifically, we first apply K-means clustering across all sentences from all reports to obtain initial cluster centroids. Then, for each report, we fit a Gaussian GMM initialized with the K-means centroids on the sentence embeddings to learn representative clusters. The means of the Gaussian components serve as the local representations of the clinical report, denoted as \( X_{RPT}^{L} \), where $X_{RPT}^{L} = \{ x_{j} \mid j = 1, \dots, C \}, \quad x_{j} \in \mathbb{R}^{D}$. We set $C$ to 16.
% Here, \( x_{j} \) represents the mean of the \( j \)-th Gaussian component, capturing the diverse semantic structures and key contextual patterns in the given clinical report. 
% By leveraging GMM-based clustering, our approach effectively models both the global meaning and the fine-grained variations present in textual healthcare data, ensuring a robust and interpretable representation for downstream analysis.

\subsection{Adaptive Cross-Modal Attention}

To effectively integrate feature representations across modalities at the local and global levels, we adopt the Cross-Modal Attention (CMA) mechanism inspired by \cite{wei2024mecom}, as illustrated in Fig. \ref{fig:model} (j). CMA allows one modality to guide attention towards the most relevant features in another, enhancing multimodal fusion.

Given three modalities—WSIs (\( X^{G}_{\text{WSI}} \), \( X^{L}_{\text{WSI}} \)), RNA-Seq (\( X^{G}_{\text{RNA}} \), \( X^{L}_{\text{RNA}} \)), and clinical reports (\( X^{G}_{\text{RPT}} \), \( X^{L}_{\text{RPT}} \))—we utilize CMA to obtain the final representations at the global ($X_{agg}^{G}$) and local ($X_{agg}^{L}$) levels. The global level final representation is computed using Eqs. \ref{eq:3}-\ref{eq:8}. First, the pairwise interactions at the global level for WSIs with RNA-seq representations ($X^{G}_{\text{WSI} \leftrightarrow \text{RNA}}$) and clinical reports ($X_{\text{WSI} \leftrightarrow \text{RPT}}^{G}$) are computed using Eqs \ref{eq:3}-\ref{eq:4}.
\begin{equation}
X^{G}_{\text{WSI} \leftrightarrow \text{RNA}} = X^{G}_{\text{WSI}} + ((X_{\text{WSI}}^{G})^{\top} X_{\text{RNA}}^{G}) \cdot X_{\text{RNA}}^{G},
\label{eq:3}
\end{equation}
\begin{equation}
X_{\text{WSI} \leftrightarrow \text{RPT}}^{G} = X_{\text{WSI}}^{G} + ((X_{\text{WSI}}^{G})^\top X_{\text{RPT}}^{G}) \cdot X_{\text{RPT}}^{G}.
\label{eq:4}
\end{equation}
Similarly, interactions for RNA-Seq and clinical reports are computed. The final aggregated representation ($X_{\text{agg}}^{G}$) is obtained using Eqs \ref{eq:5}-\ref{eq:8} follows:
\begin{equation}
X_{\text{WSI}}^{\text{CMA\_G}} = \frac{X_{\text{WSI} \leftrightarrow \text{RNA}}^{G} + X_{\text{WSI} \leftrightarrow \text{RPT}}}{2},
\label{eq:5}
\end{equation}
\begin{equation}
X_{\text{RNA}}^{\text{CMA\_G}} = \frac{X_{\text{RNA} \leftrightarrow \text{WSI}} + X_{\text{RNA} \leftrightarrow \text{RPT}}}{2}.
\label{eq:6}
\end{equation}
\begin{equation}
X_{\text{RPT}}^{\text{CMA\_G}} = \frac{X_{\text{RPT} \leftrightarrow \text{WSI}} + X_{\text{RPT} \leftrightarrow \text{RNA}}}{2}.
\label{eq:7}
\end{equation}
\begin{equation}
X_{\text{agg}^{G}} = X_{\text{WSI}}^{\text{CMA\_G}} + X_{\text{RNA}}^{\text{CMA\_G}} + X_{\text{RPT}}^{\text{CMA\_G}}
\label{eq:8}
\end{equation}
Following \cite{wei2024mecom}, we compute the final multimodal representation at the local level using pairwise interactions between key local representations of modalities using the same methodology as for the global level via Eqs. \ref{eq:3}-\ref{eq:8} (see supplementary). These key local representations are the most informative ones, selectively activated through sparse Mixture of Experts (MoE) \cite{shazeer2017outrageously}.
To illustrate our approach, we describe the process applied to the local representations extracted from WSIs, as the same methodology is employed for the other modalities.
For a given WSI, the local representations is defined as \( X_{WSI}^{L} = \{x_{j} \mid j = 1, \dots, C\} \).
where \( C \) is the number of local features and sets to 16. These representations are processed by a set of $k$ expert networks \( \{EXP_k(\cdot)\}_{k=1}^{K} \), with selection weights ($\alpha_k$) assigned by a gating network ($Gate_{k}$), as described in Eq. \ref{eq:9}:
\begin{equation}
z_{j} = \sum_{k \in \mathcal{S}_{j}} Gate_{k}(x_{j}) EXP_k(x_{j})
\label{eq:9}
\end{equation}
The number of experts is set to 5. To encourage sparsity, we apply Top-K selection, retaining only the highest-scoring experts with k = 2. $\mathcal{S}_{j}$ represents the set of selected experts for a given local representation  $x_j$  after the Top-K selection process.
Then, the final activation gating function (\( Act(\cdot) \)) selects the most informative local features, and the final WSI representation is computed as:
\begin{equation}
\begin{aligned}
    X_{WSI}^{L^{*}} &= Z \odot \mathbf{1}_{\mathcal{A}}, \\
    \mathcal{A} &= \text{Top-k} (Act(z_{1}), Act(z_{2}), \dots, Act(z_{C})).
\end{aligned}
\end{equation}

This approach selects the most relevant features while suppressing redundancy. (See supplementary for details)

\subsection{Decoupled Reconstruction for Missing Modality Completion}

To effectively handle missing modalities, we adopt a Decoupled Reconstruction mechanism from \cite{wei2024mecom}, which is responsible for reconstructing the respective information of each modality, as illustrated in Fig. \ref{fig:model}l. 
Given the aggregated multimodal representations \( X_{\text{agg}}^{G} \) and \( X_{\text{agg}}^{L} \), the model decouples them into modality-specific feature vectors via a two-layer perceptron with ReLU activation \( \psi_m(\cdot) \). This modality-agnostic process enables reconstructing missing modalities independently. The reconstructed representation for WSIs at the global level is:
\begin{equation}
X_{\text{WSI}}^{\text{CMA\_RG}} = \psi_{\text{WSI}}(X_{\text{agg}}^{G})
\end{equation}
To ensure accurate decoupling, we minimize the reconstruction loss between the decoupled modality-specific feature and the original fused representation via Eq. \ref{eq:12}.
\begin{align}
\mathcal{L}_{\text{rec}}^{G} &= \| X_{\text{WSI}}^{\text{CMA\_G}} - X_{\text{WSI}}^{\text{CMA\_RG}} \|_2^2 + \notag \\
&\quad \| X_{\text{RNA}}^{\text{CMA\_G}} - X_{\text{RNA}}^{\text{CMA\_RG}} \|_2^2 + \notag \\
&\quad \| X_{\text{RPT}}^{\text{CMA\_G}} - X_{\text{RPT}}^{\text{CMA\_RG}} \|_2^2
\label{eq:12}
\end{align}
The reconstruction loss function is shown in Eq. \ref{eq:13}. where $\mathcal{L}_{\text{rec}}^{L}$ is the reconstruction loss at the local level and is computed similarly to the global reconstruction loss.

\begin{equation}
\mathcal{L}_{\text{rec}}=\mathcal{L}_{\text{rec}}^{G}+\mathcal{L}_{\text{rec}}^{L}
\label{eq:13}
\end{equation}

\subsection{Final Feature Aggregation Using MOE}

As shown in Fig. \ref{fig:model}, our model incorporates a MoE aggregation mechanism to dynamically fuse multimodal features and optimize their contribution to the final prediction. This ensures selective integration of informative features from WSIs, RNA-Seq, and clinical reports while suppressing redundancy.
The final feature representation for each modality depends on its presence ($M_{pre}=1$) or absence ($M_{pre}=0$), incorporating reconstructed features for missing modalities. The final representation for the WSI at the global level is determined using Eq \ref{eq:14}. 
\begin{equation}
X_{\text{WSI}}^{\text{Final\_G}} =M_{pre}^{WSI}X_{\text{WSI}}^{\text{CMA\_G}}+(1-M_{pre}^{WSI})X_{\text{WSI}}^{\text{CMA\_RG}} 
\label{eq:14}
\end{equation}
The final representations for the remaining modalities (\(X_{\text{WSI}}^{\text{Final\_L}}\), \(X_{\text{RNA}}^{\text{Final\_G}}\), \(X_{\text{RNA}}^{\text{Final\_L}}\), \(X_{\text{RPT}}^{\text{Final\_G}}\), and \(X_{\text{RPT}}^{\text{Final\_L}}) \) at both global and local levels are computed similarly.
% To construct the final representation ($X_{out}$), we first stack the reconstructed feature representations from different modalities:
% \begin{flalign}
% X_{\text{stacked}} = \{ &X_{\text{WSI}}^{\text{Final\_G}}, X_{\text{WSI}}^{\text{Final\_L}}, 
% X_{\text{RNA}}^{\text{Final\_G}}, X_{\text{RNA}}^{\text{Final\_L}},  & \notag \\
% &X_{\text{RPT}}^{\text{Final\_G}}, X_{\text{RPT}}^{\text{Final\_L}} \} &&
% \end{flalign}
Then the final MoE module is utilized to adaptively selects and fuses the most informative features across modalities. The expert-based fusion is performed according to Eq.~\ref{eq:9}, producing the final representation \( X_{\text{out}} \). 
A gating network assigns selection scores, and Top-K selection retains the highest-scoring experts. The final multimodal representation is a weighted sum of selected expert outputs.

\subsection{Aligning}

We employ Symmetric Contrastive Learning (SymCL) and Multi-Prototype Contrastive Learning to align multimodal representations and refine feature embeddings for precise classification.
To ensure consistency between each modality and the aggregated representation ($X_{\text{agg}}$) at local and global levels, we use a symmetric cross-modal contrastive objective (Fig.~\ref{fig:model}k). This approach maintains alignment while preserving modality-specific characteristics. SymCL, widely used in visual-language pretraining \cite{lu2023visual,radford2021learning}, enforces sample-level alignment and prevents feature collapse.
Given a batch of \( M \) WSIs, RNA-Seq, and clinical reports, the contrastive objective for aligning WSI global representation ($X_{\text{WSI}}^{\text{CMA\_G}}$) with the global aggregated representation ($X_{\text{agg}}^{G}$) using Eq. \ref{eq:15}:
\begin{align}
\mathcal{L}_{\text{SymCL}}^{WSI,G} = & - \frac{1}{2M} \sum_{i=1}^{M} \log \frac{\exp{(\tau (X_{WSI}^{CMA\_G})_i^{T} X_{agg_{i}}^{G}})}{\sum_{j=1}^{M} \exp{(\tau (X_{WSI}^{CMA\_G})_i^{T} X_{agg_{j}}^{G}})} \notag \\
& - \frac{1}{2M} \sum_{j=1}^{M} \log \frac{\exp{(\tau (X_{agg}^{G})^{T}_{j} X_{WSI_{j}}^{CMA\_G}})}{\sum_{i=1}^{M} \exp{(\tau (X_{agg}^{G})^{T}_{j} X_{WSI_{i}}^{CMA\_G}})}
\label{eq:15}
\end{align}
The first term aligns WSI representations with their corresponding aggregated representations, while the second term enforces the reverse alignment. This bidirectional alignment ensures robustness and flexibility, preventing feature collapse, while allowing each sample to maintain a fine-tuned representation.
The symmetric contrastive losses for other modalities at global and local levels follow the same formulation, with the final loss given by Eq.\ref{eq:16}.
\begin{equation}
\begin{split}
\mathcal{L}_{\text{SymCL}} = [\mathcal{L}_{\text{SymCL}}^{WSI,G} + \mathcal{L}_{\text{SymCL}}^{RNA,G} + \mathcal{L}_{\text{SymCL}}^{RPT,G} \\
+ \mathcal{L}_{\text{SymCL}}^{WSI,L} + \mathcal{L}_{\text{SymCL}}^{RNA,L} + \mathcal{L}_{\text{SymCL}}^{RPT,L}]/6
\label{eq:16}
\end{split}
\end{equation}

We utilize Multi-Prototype Contrastive Learning \cite{moradinasab2024gengmm,moradinasab2024protogmm} to enhance MoSARe by modeling the feature distribution \( X_{\text{out}} \) with a Gaussian Mixture Model (GMM) per class, parameterized as \( \phi_c = \{ \pi_c, \mu_c, \Sigma_c \} \), where \( \pi_c \) denotes mixture weights, and \( \mu_c, \Sigma_c \) are the mean and covariance. These parameters are optimized online using a momentum-based Sinkhorn Expectation-Maximization (EM) algorithm \cite{liang2022gmmseg}. The feature distribution is modeled as $p(X_{out} | c; \phi_c) = \sum_{m=1}^{M} \pi_{c,m} \mathcal{N} (X_{out}; \mu_{c,m}, \Sigma_{c,m})$, where the constraint \( \sum_{m=1}^{M} \pi_{c,m} = 1 \) holds.
Contrastive learning aligns feature embeddings \( X_{\text{out}} \) with the most representative GMM components via:
\begin{equation}
\mathcal{L}_{\text{MCL}} = - \log \frac{\exp(X_{\text{out}}^{T} q^+ / \tau)}{\exp(X_{\text{out}}^{T} q^+ / \tau) + \sum_{c' \neq c} \exp(X_{\text{out}}^{T} q^-_{c'} / \tau)}.
\end{equation}
\begin{equation}
q^+ = \{\mu_{c,m^+}|\quad m^+ = \arg \max_{m} p(m | X_{out}, c; \phi_c)\}
\end{equation}
\begin{equation}
q^-_{c'}=\{\mu_{{c'},m^-}, \quad m^- = \arg \max_{m} p(m | X_{out}, c'; \phi_{c'})\}
\end{equation}
where  \( q^+ \)  is the positive prototype—the closest GMM component within the same class ($c$)—and  \( q^- \)  is the negative prototype, which is the closest component from a different class (i.e. $c^{'}\neq c$). 
This approach improves representation learning by aligning intra-class features while maintaining inter-class separation. Furthermore, $p(m | X_{out}, c; \phi_{c})$ represents the posterior probability of the GMM. Both symmetric contrastive learning and multi-prototype contrastive learning will be performed at each epoch after the warm-up epochs which is set to 10.

\subsection{Loss Function}

In our model, the total loss function comprises four key components: \textit{symmetric contrastive loss}, \textit{multi-prototype contrastive loss}, \textit{reconstruction loss}, and \textit{classification loss}, ensuring effective multimodal representation learning. The classification loss consists of global ($\mathcal{L}_{\text{cls}}^{\text{G}}$), local ($\mathcal{L}_{\text{cls}}^{\text{L}}$), and aggregated terms ($\mathcal{L}_{\text{cls}}^{\text{agg}}$), applied to each modality’s final representation and the multimodal output \( X_{\text{out}} \). The total loss function is formulated in Eq. \ref{eq:20}. The hyperparameters \( \lambda_1, \lambda_2, \lambda_3, \lambda_4 \) control the contribution of each loss term. 
\begin{equation}
\mathcal{L}_{\text{total}} = \lambda_1 \mathcal{L}_{\text{SymCL}} + \lambda_2 \mathcal{L}_{\text{MCL}} + \lambda_3 
\mathcal{L}_{\text{rec}}+ \lambda_4 (\mathcal{L}_{\text{cls}}^{\text{G}} + \mathcal{L}_{\text{cls}}^{\text{L}} + \mathcal{L}_{\text{cls}}^{\text{agg}})
\label{eq:20}
\end{equation}

% where \( \mathcal{L}_{\text{SymCL}} \) ensures alignment between modalities and their aggregated representation, \( \mathcal{L}_{\text{GMMCL}} \) is the multi-prototype contrastive loss leveraging GMM-based alignment, \( \mathcal{L}_{\text{rec}} \) reconstructs missing modalities, and \( \mathcal{L}_{\text{cls}}^{\text{G}}, \mathcal{L}_{\text{cls}}^{\text{L}}, \mathcal{L}_{\text{cls}}^{\text{agg}} \) correspond to classification losses at the global, local, and aggregated levels, respectively. 

% ensuring balanced learning across multimodal feature spaces while maintaining robust classification performance. 
% This formulation enforces consistency between different modalities while optimizing feature representations for downstream predictive tasks.

\begin{table*}
\setlength{\tabcolsep}{1pt} % Adjust column spacing
  \centering
  \begin{tabular}{@{}l|c|c|c|c|c|c|c|c|c@{}}
    \toprule
    \multirow{2}{*}{\textbf{Method}} 
    & \multicolumn{3}{c|}{\textbf{TCGA-BRCA}} 
    & \multicolumn{3}{c|}{\textbf{TCGA-RCC}} 
    & \multicolumn{3}{c}{\textbf{TCGA-NSCLC}} \\
    \cmidrule(lr){2-4} \cmidrule(lr){5-7} \cmidrule(lr){8-10}
    & AUC  & F1-Score & Acc 
    & AUC  & F1-Score & Acc 
    & AUC  & F1-Score & Acc \\
    \midrule
    SNN (omics)\cite{klambauer2017snn} & \(\mathbf{92.50\!\pm\! 0.8}\) & \(69.57\! \pm\! 2.3\) & \(87.60\! \pm \!3.6\) & \(\underline{98.84\! \pm \!0.8}\) & \(93.25\! \pm \!2.8\) & \(95.41\! \pm \!1.9\) & \(\underline{97.33 \!\pm \!1.2}\) & \(92.09\! \pm \!2.6\) &\(92.21 \!\pm \!2.5\)\\
    ABMIL(WSI)\cite{ilse2018attention} & \(85.88\!\pm\! 7.1\) & \(79.74\!\pm\! 10.0\) & \(\underline{93.48\!\pm\! 2.9}\) & \(96.11\!\pm\! 3.1\) & \(\underline{95.46\!\pm\! 3.7}\) & \(\underline{96.05\!\pm\! 3.2}\) & \(94.03\!\pm\! 2.7\) & \(94.06\!\pm\! 2.7\) & \(94.03\!\pm\! 2.7\) \\ 
    GSCNN\cite{mobadersany2018GSCNN}  & \(78.45\! \pm \!6.4\) & \(69.06\! \pm \!9.2\) & \(89.03\! \pm \!3.1\) & \(69.1\! \pm \!1.0\) & \(89.03\! \pm \!3.9\) & \(94.33\! \pm \!3.1\) & \(91.44\! \pm \!2.6\) & \(91.07 \!\pm\! 2.8\) &\(91.17 \!\pm\! 2.5\) \\ 
    BP\cite{chen2022pathomicbp}  & \( 80.00\!\pm \!5.0\) & \(72.53\! \pm \!6.6\) &\(89.62\! \pm \!2.7\)  & \(95.30 \!\pm\! 2.9\) & \( 94.41\!\pm \!3.7\) & \( 95.10\!\pm \!2.9\) & \(92.31\! \pm\! 2.0\) & \(92.16 \!\pm\! 3.05\) & \( 92.55\!\pm \!1.5\)\\ 
    MCAT\cite{chen2021multimodal}  & \(81.43 \!\pm \!7.1\) & \(72.59 \!\pm \!10.8\) & \(91.13\! \pm \!3.8\) & \(98.83\! \pm \!2.3\) & \(95.31 \!\pm \!2.6\) &\(96.02\! \pm\! 2.1\)&\(97.10\! \pm \!1.5\) & \(\underline{97.10\! \pm\! 1.5}\) & \( \underline{97.10\!\pm \!1.5}\)\\ 
    PORPOISE\cite{chen2022pan}  & \(88.43\! \pm\! 4.7\) & \(\underline{81.45\!\pm \!6.1}\) & \(\mathbf{93.48\!\pm\! 2.3}\) & \(96.01 \!\pm\! 3.1\) & \(95.42 \!\pm \!3.7\) & \(96.03 \!\pm\! 3.1\) & \(93.91\! \pm\! 2.9\) & \(93.96 \!\pm \!2.8\) &\(93.92 \!\pm\! 2.9\)\\ 
    MoSARe(ours) & \( \underline{90.66\!\pm \!2.1}\)  & \( \mathbf{81.48\!\pm \!4.1}\)  & \( 91.89\!\pm\! 1.9\)  & \(\mathbf{99.53\!\pm\!0.8}\) & \(\mathbf{99.29\!\pm\!1.2}\) & \(\mathbf{99.50\!\pm\!0.8}\) & \( \mathbf{98.66\!\pm\! 0.8}\) & \( \mathbf{98.65\!\pm \!1.0}\) & \( \mathbf{98.65\!\pm\! 0.9}\) \\ 
    \bottomrule
  \end{tabular}
  \caption{5-fold CV classification results of different algorithms on TCGA-BRCA, TCGA-RCC, and TCGA-NSCLC datasets. Only using images and genomic data. Highest performing model is bold and the second highest underlined.}
  \label{tab:comparison_results}
\end{table*}

\begin{table*}[ht]
    \centering
    \setlength{\tabcolsep}{2pt} % Adjust column spacing
    \renewcommand{\arraystretch}{1.1} % Adjust row spacing
    \begin{tabular}{@{}c|l|l|ccc|ccc|ccc@{}}
        \toprule
        \multirow{2}{*}{\textbf{Masking \%}} & \multirow{2}{*}{\textbf{Train}} & \multirow{2}{*}{\textbf{Test}} & 
        \multicolumn{3}{c|}{\textbf{TCGA-BRCA}} & \multicolumn{3}{c|}{\textbf{TCGA-RCC}} & \multicolumn{3}{c}{\textbf{TCGA-NSCLC}} \\
        \cmidrule(lr){4-6} \cmidrule(lr){7-9} \cmidrule(lr){10-12}
        & & & \textbf{AUC} & \textbf{F1-Score} & \textbf{Acc} & \textbf{AUC} & \textbf{F1-Score} & \textbf{Acc} & \textbf{AUC} & \textbf{F1-Score} & \textbf{Acc} \\
        \midrule
        \multirow{3}{*}{10\%}  
        & Masked    & Masked   & 87.80 & 78.78 & 91.12 & 97.41 & 96.57 & 97.50 & 97.25 & 97.27 & 97.24 \\
        & Masked   & Unmasked & 89.03 & 81.55 & 92.45 & 99.20 & 98.94 & 99.25 & 98.45 & 98.48 & 98.47 \\
        & Removed & Unmasked & 90.17 & 81.54 & 91.98 & 98.95 & 98.47 & 98.89 & 97.84 & 97.30 & 97.25 \\
        \midrule
        \multirow{3}{*}{30\%}  
        & Masked   & Masked   & 87.04 & 77.35 & 90.46 & 96.30 & 95.06 & 96.40 & 95.69 & 95.72 & 95.69 \\
        & Masked   & Unmasked & 89.13 & 81.51 & 92.36 & 99.35 & 99.26 & 99.49 & 98.31 & 98.31 & 98.29 \\
        & Removed & Unmasked & 85.22 & 70.22 & 85.18 & 90.83 & 87.34 & 89.28 & 56.96 & 23.35 & 56.18 \\
        \midrule
        \multirow{3}{*}{50\%}  
        & Masked  & Masked  &  85.72 & 76.19 & 90.04  & 93.06 & 91.77 & 94.22 & 90.12 & 90.50 & 90.18 \\
        & Masked  & Unmasked & 89.43 & 77.50 & 89.28 & 96.72 & 95.74 & 97.02 & 94.15 & 94.54 & 94.52 \\
        & Removed & Unmasked & NA & NA & NA & NA & NA & NA & NA & NA & NA \\
        \bottomrule
    \end{tabular}
    \caption{Evaluation of Different Masking Strategies in Training and Testing across TCGA Datasets. NA=Not Applicable}
    \label{tab:masking_experiment_all}
\end{table*}

\begin{table}[ht]
  \centering
  \setlength{\tabcolsep}{1pt} % Adjust column spacing
  \begin{tabular}{@{}c|c|c|c|c@{}}
    \toprule
    \textbf{Dataset} & \textbf{Modality} & \textbf{AUC} & \textbf{F1-Score} & \textbf{Acc} \\
    \midrule
    \multirow{2}{*}{TCGA-BRCA} & 2 Modality & 88.78 & 75.94 & 88.11 \\
    & 3 Modality & 87.95 & 79.61 & 91.54 \\
    \midrule
    \multirow{2}{*}{TCGA-RCC} & 2 Modality & 99.70 & 99.52 & 99.63 \\
    & 3 Modality & 99.82 & 99.82 & 99.87 \\
    \midrule
    \multirow{2}{*}{TCGA-NSCLC} & 2 Modality & 98.78 & 98.82 & 98.78 \\
    & 3 Modality & 99.08 & 99.12  & 99.08 \ \\
    \bottomrule
  \end{tabular}
  \caption{Comparison of Results on Two vs. Three Modalities}
  \label{tab:performance-tree_modality}
\end{table}

\begin{table}[ht]
  \centering
  \begin{tabular}{@{}c|c|c|c@{}}
    \toprule
    \textbf{CMA} & \textbf{MOE} & \textbf{Aligning} & \textbf{AUC} \\
    \midrule
    \xmark & \xmark & \xmark & 95.68 \\
    \cmark & \cmark & \xmark & 98.37 \textcolor{red}{(\textbf{+2.69} \(\uparrow\))} \\
    \cmark & \cmark & \cmark & 99.08 \textcolor{red}{(\textbf{+3.4} \(\uparrow\))} \\
    \bottomrule
  \end{tabular}
  \caption{Ablation study on the complete dataset (3 Modalities).}
  \label{tab:ablation_study_3modality}
\end{table}

\begin{table}[ht]
  \centering
  \begin{tabular}{@{}c|c|c|c|c@{}}
    \toprule
    \textbf{CMA} & \textbf{Recons} & \textbf{MOE} & \textbf{Aligning} & \textbf{AUC} \\
    \midrule
    \cmark & \xmark & \xmark & \xmark & 93.16 \\
    \cmark & \cmark & \xmark & \xmark & 93.82 \textcolor{red}{(\textbf{+0.66} \(\uparrow\))} \\
    \cmark & \cmark & \cmark & \xmark & 94.03 \textcolor{red}{(\textbf{+0.87} \(\uparrow\))} \\
    \cmark & \cmark & \cmark & \cmark & 95.69 \textcolor{red}{(\textbf{+2.53} \(\uparrow\))} \\
    \bottomrule
  \end{tabular}
  \caption{Ablation study on the incomplete dataset (2 Modalities).}
  \label{tab:ablation_study_2modality}
\end{table}
\section{Experiment}
\label{sec:results}

\textbf{Implementation Details: }
For all datasets, we extract feature representations for each modality following the preprocessing steps in Section 3.1. For WSIs, we use a pretrained UNI model \cite{chen2024towards} to obtain patch-level feature embeddings. The MoSARe model receives 2048 sampled patch embeddings per slide along with the means of GMM components as inputs, normalizes them, and applies ABMIL to the sampled patch embeddings to derive a global representation. For gene expression data, we follow the preprocessing method in \cite{jaume2024transcriptomics, jaume2024modeling}. The preprocessed data is provided as input to the model, normalized, and processed using multi-head attention and an MLP to generate both global and local representations. For clinical reports, we remove labels to prevent leakage and fine-tune a sentence transformer based on ModernBERT \cite{modernbert} to extract feature representations at both the sentence and report levels. The report-level embedding serves as a global representation, while we apply a GMM to the sentence-level embeddings to obtain local representations for the report. Our model takes these global and local features as input, normalizes them, and refines them for further processing. Regarding the hyperparameters, we set \( \lambda_{i} \) in Eq. \ref{eq:20} to 1 except \( \lambda_{2} \) which is set to 2. The network is trained using the Adam optimizer with a learning rate of \( 10^{-4} \) and a batch size of 32 for all datasets for 100 epochs. Each experiment is conducted using five-fold cross-validation to report the average results. All models are executed on an A100 GPU. Furthermore, to simulate missing modalities, we randomly mask \( n\% \) of each modality during training/testing. 
\begin{equation}
X_{\text{masked}} = X \odot M,
\label{eq:21}
\end{equation}
where \( X \) represents the original modality features, and \( M \) is a binary mask (0 for masked modalities, 1 otherwise).

\textbf{Datasets and Evaluation Metrics: }
To assess the performance of our MoSARe model, we conducted experiments on three publicly available datasets from The Cancer Genome Atlas (TCGA) from the Genomic Data Commons (GDC) portal \cite{weinstein2013cancer}.
% \footnote{\url{https://portal.gdc.cancer.gov/}}. 
These datasets cover distinct cancer subtyping tasks: (1) classification of invasive breast carcinoma (BRCA) into Invasive Ductal Carcinoma (IDC) and Invasive Lobular Carcinoma (ILC), (2) differentiation of non-small cell lung carcinoma (NSCLC) into Lung Adenocarcinoma (LUAD) and Lung Squamous Cell Carcinoma (LUSC), and (3) subtyping of Renal Cell Carcinoma (RCC) into Kidney Chromophobe (KICH), Kidney Renal Clear Cell Carcinoma (KIRC), and Kidney Renal Papillary Cell Carcinoma (KIRP). We downloaded RNA-Seq data from the Xena database \cite{goldman2020visualizing}.
% We also gathered RNA-Seq abundance from cBioPortal \footnote{\url{https://www.cbioportal.org/}} for all three datasets. 
For clinical reports associated with each patient, we used the cleaned TCGA Reports data from \citep{kefeli2024tcga}.
% available here \url{https://data.mendeley.com/datasets/hyg5xkznpx/1}.
To ensure fair comparisons with state-of-the-art (SOTA) models across all experiments, we included only patients with both WSI and RNA-Seq data for experiments involving two modalities. Similarly, for experiments incorporating three modalities (WSI, RNA-Seq, and clinical reports), we considered only patients with all three data types available. Data statistics for two and three modalities are presented in the supplementary material (Tab. S1). The table includes the distribution of WSIs and patients, as the distribution of reports closely follows that of WSIs, and RNA-Seq aligns similarly with WSIs.
For all experiments, we evaluate performance using accuracy, F1-score, and AUC (Area Under the Curve). 
% Accuracy measures how many samples are correctly classified. F1-score balances precision and recall, making it useful for imbalanced data. AUC shows how well the model distinguishes between classes, with higher values indicating better performance.

\subsection{Results and Analysis} 
In this section, we provide the implementation details, empirical settings, main results, and ablation studies.

\textbf{Comparisons with State-of-the-art Methods:} We compare the performance of MoSARe with the following state-of-the-art (SOTA) models:
\begin{itemize}
    \item \textbf{SNN}\cite{klambauer2017snn}: This unimodal genomic baseline enhances MLPs by replacing ReLU with self-normalizing SELU activations, using Alpha Dropout for variance preservation, and applying LeCun initialization.
    \item \textbf{ABMIL}\cite{ilse2018attention}: This Multiple Instance Learning (MIL) method uses attention based pooling to select for relevant WSI patches and use them for classification.
    % \textcolor{red}{(Saurav’s input)}
    \item \textbf{BP}\cite{chen2022pathomicbp}: A multimodal fusion algorithm uses Kronecker Product to combine histology and genomic features.
    \item \textbf{PORPOISE}\cite{chen2022pan}: A multimodal model, using imaging and genomic data built by combining ABMIL and SNN.
    % \textcolor{red}{(Saurav’s input)}
    \item \textbf{GSCNN}\cite{mobadersany2018GSCNN}: A strategy for combining histology image and genomic features via vector concatenation.
    \item \textbf{MCAT}\cite{chen2021multimodal}: A multimodal model that uses uses co-attention to learn pairwise interactions between imaging and genomic data and use it for risk prediction.
    % \textcolor{red}{(Saurav’s input)}
\end{itemize}
To keep the comparison fair, all methods used the same 5-fold cross-validation splits and trained over two modalities WSIs and RNA-seq data per dataset. Each reference method was tested with its original code and settings. Furthermore, all experiments have been conducted only on two classes of the TCGA-RCC dataset, as the preprocessed RNA-seq data for some of the SOTA methods were available only for these two classes, specifically for PORPOISE and MCAT \cite{chen2021multimodal, chen2022pan}.
% \textcolor{red}{(Saurav’s input: Reference should be included)}. 
Tab. \ref{tab:comparison_results} shows the classification results of various algorithms on the TCGA-BRCA, TCGA-NSCLC, and TCGA-RCC datasets. For the imbalanced BRCA dataset, MoSARe achieves an AUC of 90.66 ± 2.1, an F1-score of 81.48 ± 4.1, and an accuracy of 91.89 ± 1.99. While SNN attains a slightly higher AUC (92.50 ± 0.79), MoSARe demonstrates substantial improvements in F1-score (81.48 vs. 69.57) and accuracy (91.89 vs. 87.60), highlighting its superior classification performance in these aspects. Additionally, PORPOISE achieves a marginally higher accuracy (93.48 ± 2.3); however, MoSARe outperforms it in AUC and F1-score, indicating a more balanced classification capability. These results suggest that MoSARe provides a robust and well-rounded performance across multiple evaluation metrics. On the TCGA-RCC dataset, MoSARe achieves the best results, with an AUC of 99.53 ± 0.8, an F1-score of 99.29 ± 1.2, and an accuracy of 99.50 ± 0.8, improving by at least 0.7\%, 4.06\%, and 3.61\%, respectively. Similarly, on the TCGA-NSCLC dataset, MoSARe attains a mean AUC of 98.66, and F1-score of 98.65, and an accuracy of 98.65, outperforming other algorithms by at least 1.37\%, 1.60\%, and 1.60 \%,  respectively. These results demonstrate MoSARe’s superior performance across all datasets.

\textbf{Comparison Results on Missing Modalities:} 	We evaluated our model’s ability to handle missing modalities through experiments with 10\%, 30\%, and 50\% missing data across three training and testing scenarios:  
  
\begin{itemize}
    \item \textbf{Masked Train - Masked Test:} Both training and testing sets have missing modalities, simulating real-world conditions with incomplete data during training and test.
    \item \textbf{Masked Train - Unmasked Test:} The training set contains missing modalities, while the test set is complete. This scenario assesses the model's ability to generalize from incomplete training data to fully available test data.
    \item \textbf{Removed Train - Unmasked Test:} Instances with missing modalities are removed from the training set, ensuring that only complete samples are used for training. The test set remains complete. This approach is commonly used in multimodal learning, as many existing models do not inherently support missing modalities.
\end{itemize}

Tab. \ref{tab:masking_experiment_all} demonstrates that the training MoSARe on incomplete data (Scenario 2: Masked Train - Unmasked Test) consistently outperforms the approach of removing incomplete samples (Scenario 3: Removed Train - Unmasked Test) across all missing modality levels (10\%, 30\%, and 50\%) on all three datasets,
% (i.e., TCGA-BRCA, TCGA-RCC, and TCGA-NSCLC),
except in the TCGA-BRCA 10\% scenario, where the performance is comparable. At every masking percentage, Scenario 2 achieves higher AUC, F1-score, and accuracy, indicating that incomplete data should not be discarded, as doing so leads to a significant loss of useful information. This highlights MoSARe’s ability to effectively learn from incomplete modalities, leveraging the available partial data to improve classification performance. The `NA' in the tables indicates that we could not train the model in Scenario 3: Removed Train - Unmasked Test (50\%) because the entire training set contains missing modalities, and no complete data is available for training. Furthermore, the model performs well even when tested on incomplete data (Scenario 1: Masked Train - Masked Test), proving its capability to generalize under real-world conditions where missing modalities are present in both training and inference stages. These findings confirm that MoSARe is highly robust in handling missing modalities, making it a powerful solution for multimodal learning tasks. 

% \vspace{-0.3cm}
\textbf{Comparison of Results on Two vs. Three Modalities:}  
Tab. \ref{tab:performance-tree_modality} presents the comparison results between two vs. three modalities across all three datasets. In the three-modality setup, we included text as the third modality. As observed, the MoSARe model trained using all three modalities achieves slightly higher performance in terms of AUC, F1-Score, and Accuracy on both TCGA-RCC and TCGA-NSCLC datasets. Furthermore, it noticeably improves the F1-score and accuracy on the highly imbalanced TCGA-BRCA dataset, even though the AUC dropped slightly. These results show the advantages of including text as the third modality, as it provides additional information.

% \vspace{-0.3cm}
\textbf{Ablation Study:}
To evaluate the effectiveness of each component in MoSARe, we systematically removed individual components and ran the model with 3 modalities and 2 modalities on both complete and incomplete TCGA-NSCLC datasets. Tab. \ref{tab:ablation_study_3modality} highlights the positive contribution of each component— Cross-Modal Attention (CMA), MOE, and the Aligning components—when the model was trained on the complete TCGA-NSCLC dataset with 3 modalities. We considered the contributions of MOE and CMA together, as MOE is part of CMA, such as preparing the local representation input to CMA. Since the dataset is complete, we did not include the reconstruction contribution. Tab. \ref{tab:ablation_study_2modality} similarly shows in detail the contribution of various parts of the model when trained with 2 modalities and 30\% missing data on TCGA-NSCLC dataset.

\textbf{Interpretability:}
We are able to generate heatmaps based on the cross-attention values, which can be a useful interpretation tool for clinicians evaluating this model. You can find the generated heatmaps and highlighted text reports for the local clusters with the highest cross-modal attention score in Supplementary materials.

\vspace{-0.35cm}
\section{Conclusion and limitations}
\label{sec:conclusion}
In this work, we introduced MoSARe, a novel model designed to address the challenges of incomplete multimodal data in healthcare applications. Our model integrates MOE, aligning, and reconstruction to enhance feature representation, improve modality consistency, and enable robust decision-making even when certain modalities are missing. Unlike existing models that assume complete data availability, MoSARe effectively handles missing modalities without relying on strong imputation assumptions. Our method outperforms other SOTAs in three datasets, including TCGA-BRCA, TCGA-RCC, and TCGA-NSCLC. Limitations include fixing  C = 16  (number of components) for all tasks, which may lead to suboptimal solutions. Future work will investigate the impact of varying C.

{
    \small
    \bibliographystyle{ieeenat_fullname}
    \bibliography{main}

\begin{thebibliography}{34}
\providecommand{\natexlab}[1]{#1}
\providecommand{\url}[1]{\texttt{#1}}
\expandafter\ifx\csname urlstyle\endcsname\relax
  \providecommand{\doi}[1]{doi: #1}\else
  \providecommand{\doi}{doi: \begingroup \urlstyle{rm}\Url}\fi

\bibitem[Chen et~al.(2021)Chen, Lu, Weng, Chen, Williamson, Manz, Shady, and Mahmood]{chen2021multimodal}
Richard~J Chen, Ming~Y Lu, Wei-Hung Weng, Tiffany~Y Chen, Drew~FK Williamson, Trevor Manz, Maha Shady, and Faisal Mahmood.
\newblock Multimodal co-attention transformer for survival prediction in gigapixel whole slide images.
\newblock In \emph{Proceedings of the IEEE/CVF international conference on computer vision}, pages 4015--4025, 2021.

\bibitem[Chen et~al.(2022{\natexlab{a}})Chen, Lu, Williamson, Chen, Lipkova, Noor, Shaban, Shady, Williams, Joo, et~al.]{chen2022pan}
Richard~J Chen, Ming~Y Lu, Drew~FK Williamson, Tiffany~Y Chen, Jana Lipkova, Zahra Noor, Muhammad Shaban, Maha Shady, Mane Williams, Bumjin Joo, et~al.
\newblock Pan-cancer integrative histology-genomic analysis via multimodal deep learning.
\newblock \emph{Cancer cell}, 40\penalty0 (8):\penalty0 865--878, 2022{\natexlab{a}}.

\bibitem[Chen et~al.(2024)Chen, Ding, Lu, Williamson, Jaume, Song, Chen, Zhang, Shao, Shaban, et~al.]{chen2024towards}
Richard~J Chen, Tong Ding, Ming~Y Lu, Drew~FK Williamson, Guillaume Jaume, Andrew~H Song, Bowen Chen, Andrew Zhang, Daniel Shao, Muhammad Shaban, et~al.
\newblock Towards a general-purpose foundation model for computational pathology.
\newblock \emph{Nature Medicine}, 30\penalty0 (3):\penalty0 850--862, 2024.

\bibitem[Chen et~al.(2022{\natexlab{b}})]{chen2022pathomicbp}
R.~J. Chen et~al.
\newblock Pathomic fusion: An integrated framework for fusing histopathology and genomic features for cancer diagnosis and prognosis.
\newblock \emph{IEEE Transactions on Medical Imaging}, 41\penalty0 (4):\penalty0 757--770, 2022{\natexlab{b}}.

\bibitem[Ding et~al.(2024)Ding, Li, Wang, Ying, and Shi]{ding2024multimodal}
Saisai Ding, Juncheng Li, Jun Wang, Shihui Ying, and Jun Shi.
\newblock Multimodal co-attention fusion network with online data augmentation for cancer subtype classification.
\newblock \emph{IEEE Transactions on Medical Imaging}, 2024.

\bibitem[Dosovitskiy et~al.(2020)Dosovitskiy, Beyer, Kolesnikov, Weissenborn, Zhai, Unterthiner, Dehghani, Minderer, Heigold, Gelly, et~al.]{dosovitskiy2020image}
Alexey Dosovitskiy, Lucas Beyer, Alexander Kolesnikov, Dirk Weissenborn, Xiaohua Zhai, Thomas Unterthiner, Mostafa Dehghani, Matthias Minderer, Georg Heigold, Sylvain Gelly, et~al.
\newblock An image is worth 16x16 words: Transformers for image recognition at scale.
\newblock \emph{arXiv preprint arXiv:2010.11929}, 2020.

\bibitem[Gillespie et~al.(2021)Gillespie, Jassal, Stephan, Milacic, Rothfels, Senff-Ribeiro, Griss, Sevilla, Matthews, Gong, Deng, Varusai, Ragueneau, Haider, May, Shamovsky, Weiser, Brunson, Sanati, Beckman, Shao, Fabregat, Sidiropoulos, Murillo, Viteri, Cook, Shorser, Bader, Demir, Sander, Haw, Wu, Stein, Hermjakob, and D’Eustachio]{gillespie2022reactome}
Marc Gillespie, Bijay Jassal, Ralf Stephan, Marija Milacic, Karen Rothfels, Andrea Senff-Ribeiro, Johannes Griss, Cristoffer Sevilla, Lisa Matthews, Chuqiao Gong, Chuan Deng, Thawfeek Varusai, Eliot Ragueneau, Yusra Haider, Bruce May, Veronica Shamovsky, Joel Weiser, Timothy Brunson, Nasim Sanati, Liam Beckman, Xiang Shao, Antonio Fabregat, Konstantinos Sidiropoulos, Julieth Murillo, Guilherme Viteri, Justin Cook, Solomon Shorser, Gary Bader, Emek Demir, Chris Sander, Robin Haw, Guanming Wu, Lincoln Stein, Henning Hermjakob, and Peter D’Eustachio.
\newblock The reactome pathway knowledgebase 2022.
\newblock \emph{Nucleic Acids Research}, 50\penalty0 (D1):\penalty0 D687--D692, 2021.

\bibitem[Goldman et~al.(2020)Goldman, Craft, Hastie, Repe{\v{c}}ka, McDade, Kamath, Banerjee, Luo, Rogers, Brooks, et~al.]{goldman2020visualizing}
Mary~J Goldman, Brian Craft, Mim Hastie, Kristupas Repe{\v{c}}ka, Fran McDade, Akhil Kamath, Ayan Banerjee, Yunhai Luo, Dave Rogers, Angela~N Brooks, et~al.
\newblock Visualizing and interpreting cancer genomics data via the xena platform.
\newblock \emph{Nature biotechnology}, 38\penalty0 (6):\penalty0 675--678, 2020.

\bibitem[Ilse et~al.(2018)Ilse, Tomczak, and Welling]{ilse2018attention}
Maximilian Ilse, Jakub Tomczak, and Max Welling.
\newblock Attention-based deep multiple instance learning.
\newblock In \emph{International conference on machine learning}, pages 2127--2136. PMLR, 2018.

\bibitem[Jaume et~al.(2024{\natexlab{a}})Jaume, Oldenburg, Vaidya, Chen, Williamson, Peeters, Song, and Mahmood]{jaume2024transcriptomics}
Guillaume Jaume, Lukas Oldenburg, Anurag Vaidya, Richard~J Chen, Drew~FK Williamson, Thomas Peeters, Andrew~H Song, and Faisal Mahmood.
\newblock Transcriptomics-guided slide representation learning in computational pathology.
\newblock In \emph{Proceedings of the IEEE/CVF Conference on Computer Vision and Pattern Recognition}, pages 9632--9644, 2024{\natexlab{a}}.

\bibitem[Jaume et~al.(2024{\natexlab{b}})Jaume, Vaidya, Chen, Williamson, Liang, and Mahmood]{jaume2024modeling}
Guillaume Jaume, Anurag Vaidya, Richard~J Chen, Drew~FK Williamson, Paul~Pu Liang, and Faisal Mahmood.
\newblock Modeling dense multimodal interactions between biological pathways and histology for survival prediction.
\newblock In \emph{Proceedings of the IEEE/CVF Conference on Computer Vision and Pattern Recognition}, pages 11579--11590, 2024{\natexlab{b}}.

\bibitem[Kefeli and Tatonetti(2024)]{kefeli2024tcga}
Jenna Kefeli and Nicholas Tatonetti.
\newblock Tcga-reports: A machine-readable pathology report resource for benchmarking text-based ai models.
\newblock \emph{Patterns}, 5\penalty0 (3), 2024.

\bibitem[Klambauer et~al.(2017)Klambauer, Unterthiner, Mayr, and Hochreiter]{klambauer2017snn}
G{\"u}nter Klambauer, Thomas Unterthiner, Andreas Mayr, and Sepp Hochreiter.
\newblock Self-normalizing neural networks.
\newblock In \emph{Proceedings of the 31st International Conference on Neural Information Processing Systems}, pages 972--981, Red Hook, NY, USA, 2017. Curran Associates Inc.

\bibitem[Liang et~al.(2022)Liang, Wang, Miao, and Yang]{liang2022gmmseg}
Chen Liang, Wenguan Wang, Jiaxu Miao, and Yi Yang.
\newblock Gmmseg: Gaussian mixture based generative semantic segmentation models.
\newblock \emph{Advances in Neural Information Processing Systems}, 35:\penalty0 31360--31375, 2022.

\bibitem[Liberzon et~al.(2015)Liberzon, Birger, Thorvaldsdóttir, Ghandi, Mesirov, and Tamayo]{Liberzon2015msigdb}
Arthur Liberzon, Craig Birger, Helga Thorvaldsdóttir, Mahmoud Ghandi, Jill~P. Mesirov, and Pablo Tamayo.
\newblock The molecular signatures database (msigdb) hallmark gene set collection.
\newblock \emph{Cell Systems}, 1\penalty0 (6):\penalty0 417--425, 2015.

\bibitem[Liu et~al.(2022)Liu, Chen, Jiao, Li, Sridhar, Vaidya, and Mahmood]{liu2022mgct}
Z. Liu, R.~J. Chen, N. Jiao, Z. Li, D. Sridhar, A. Vaidya, and F. Mahmood.
\newblock Mgct: Mutual-guided cross-modal transformer for survival outcome prediction using gigapixel whole slide images and genomic data.
\newblock \emph{Medical Image Analysis}, 82:\penalty0 102622, 2022.

\bibitem[Lu et~al.(2023)Lu, Chen, Zhang, Williamson, Chen, Ding, Le, Chuang, and Mahmood]{lu2023visual}
Ming~Y Lu, Bowen Chen, Andrew Zhang, Drew~FK Williamson, Richard~J Chen, Tong Ding, Long~Phi Le, Yung-Sung Chuang, and Faisal Mahmood.
\newblock Visual language pretrained multiple instance zero-shot transfer for histopathology images.
\newblock In \emph{Proceedings of the IEEE/CVF conference on computer vision and pattern recognition}, pages 19764--19775, 2023.

\bibitem[Mobadersany et~al.(2018)Mobadersany, Yousefi, Amgad, Gutman, Barnholtz-Sloan, Vega, and Cooper]{mobadersany2018GSCNN}
Pooya Mobadersany, Sahand Yousefi, Mohamed Amgad, David~A. Gutman, Jill~S. Barnholtz-Sloan, Juan E.~V. Vega, and Lee A.~D. Cooper.
\newblock Predicting cancer outcomes from histology and genomics using convolutional networks.
\newblock \emph{Proceedings of the National Academy of Sciences}, 115\penalty0 (13):\penalty0 E2970--E2979, 2018.

\bibitem[Moradinasab et~al.(2024{\natexlab{a}})Moradinasab, Jafarzadeh, and Brown]{moradinasab2024gengmm}
Nazanin Moradinasab, Hassan Jafarzadeh, and Donald~E Brown.
\newblock Gengmm: Generalized gaussian-mixture-based domain adaptation model for semantic segmentation.
\newblock In \emph{2024 IEEE International Conference on Image Processing (ICIP)}, pages 1078--1084. IEEE, 2024{\natexlab{a}}.

\bibitem[Moradinasab et~al.(2024{\natexlab{b}})Moradinasab, Shankman, Deaton, Owens, and Brown]{moradinasab2024protogmm}
Nazanin Moradinasab, Laura~S Shankman, Rebecca~A Deaton, Gary~K Owens, and Donald~E Brown.
\newblock Protogmm: Multi-prototype gaussian-mixture-based domain adaptation model for semantic segmentation.
\newblock \emph{arXiv preprint arXiv:2406.19225}, 2024{\natexlab{b}}.

\bibitem[Oquab et~al.(2023)Oquab, Darcet, Moutakanni, Vo, Szafraniec, Khalidov, Fernandez, Haziza, Massa, El-Nouby, et~al.]{oquab2023dinov2}
Maxime Oquab, Timoth{\'e}e Darcet, Th{\'e}o Moutakanni, Huy Vo, Marc Szafraniec, Vasil Khalidov, Pierre Fernandez, Daniel Haziza, Francisco Massa, Alaaeldin El-Nouby, et~al.
\newblock Dinov2: Learning robust visual features without supervision.
\newblock \emph{arXiv preprint arXiv:2304.07193}, 2023.

\bibitem[Radford et~al.(2021)Radford, Kim, Hallacy, Ramesh, Goh, Agarwal, Sastry, Askell, Mishkin, Clark, et~al.]{radford2021learning}
Alec Radford, Jong~Wook Kim, Chris Hallacy, Aditya Ramesh, Gabriel Goh, Sandhini Agarwal, Girish Sastry, Amanda Askell, Pamela Mishkin, Jack Clark, et~al.
\newblock Learning transferable visual models from natural language supervision.
\newblock In \emph{International conference on machine learning}, pages 8748--8763. PmLR, 2021.

\bibitem[Shang et~al.(2017)Shang, Palmer, Sun, Chen, Lu, and Bi]{shang2017vigan}
Chao Shang, Aaron Palmer, Jiangwen Sun, Ko-Shin Chen, Jin Lu, and Jinbo Bi.
\newblock Vigan: Missing view imputation with generative adversarial networks.
\newblock In \emph{2017 IEEE International conference on big data (Big Data)}, pages 766--775. IEEE, 2017.

\bibitem[Shazeer et~al.(2017)Shazeer, Mirhoseini, Maziarz, Davis, Le, Hinton, and Dean]{shazeer2017outrageously}
Noam Shazeer, Azalia Mirhoseini, Krzysztof Maziarz, Andy Davis, Quoc Le, Geoffrey Hinton, and Jeff Dean.
\newblock Outrageously large neural networks: The sparsely-gated mixture-of-experts layer.
\newblock \emph{arXiv preprint arXiv:1701.06538}, 2017.

\bibitem[Subramanian et~al.(2005)Subramanian, Tamayo, Mootha, Mukherjee, Ebert, Gillette, Paulovich, Pomeroy, Golub, Lander, and Mesirov]{Subramanian2005genome}
Aravind Subramanian, Pablo Tamayo, Vamsi~K. Mootha, Sayan Mukherjee, Benjamin~L. Ebert, Michael~A. Gillette, Amanda Paulovich, Scott~L. Pomeroy, Todd~R. Golub, Eric~S. Lander, and Jill~P. Mesirov.
\newblock Gene set enrichment analysis: A knowledge-based approach for interpreting genome-wide expression profiles.
\newblock \emph{Proceedings of the National Academy of Sciences}, 102\penalty0 (43):\penalty0 15545--15550, 2005.

\bibitem[Warner et~al.(2024)Warner, Chaffin, Clavié, Weller, Hallström, Taghadouini, Gallagher, Biswas, Ladhak, Aarsen, Cooper, Adams, Howard, and Poli]{modernbert}
Benjamin Warner, Antoine Chaffin, Benjamin Clavié, Orion Weller, Oskar Hallström, Said Taghadouini, Alexis Gallagher, Raja Biswas, Faisal Ladhak, Tom Aarsen, Nathan Cooper, Griffin Adams, Jeremy Howard, and Iacopo Poli.
\newblock Smarter, better, faster, longer: A modern bidirectional encoder for fast, memory efficient, and long context finetuning and inference, 2024.

\bibitem[Wei et~al.(2024)Wei, Yu, Xu, Zhang, and Peng]{wei2024mecom}
Xiu-Shen Wei, Hong-Tao Yu, Anqi Xu, Faen Zhang, and Yuxin Peng.
\newblock Mecom: A meta-completion network for fine-grained recognition with incomplete multi-modalities.
\newblock \emph{IEEE Transactions on Image Processing}, 2024.

\bibitem[Weinstein et~al.(2013)Weinstein, Collisson, Mills, Shaw, Ozenberger, Ellrott, Shmulevich, Sander, and Stuart]{weinstein2013cancer}
John~N Weinstein, Eric~A Collisson, Gordon~B Mills, Kenna~R Shaw, Brad~A Ozenberger, Kyle Ellrott, Ilya Shmulevich, Chris Sander, and Joshua~M Stuart.
\newblock The cancer genome atlas pan-cancer analysis project.
\newblock \emph{Nature genetics}, 45\penalty0 (10):\penalty0 1113--1120, 2013.

\bibitem[Wu et~al.(2024{\natexlab{a}})Wu, Wang, Chen, and Carneiro]{wu2024deep}
Renjie Wu, Hu Wang, Hsiang-Ting Chen, and Gustavo Carneiro.
\newblock Deep multimodal learning with missing modality: A survey.
\newblock \emph{arXiv preprint arXiv:2409.07825}, 2024{\natexlab{a}}.

\bibitem[Wu et~al.(2024{\natexlab{b}})Wu, Dadu, Tustison, Avants, Nalls, Sun, and Faghri]{wu2024multimodal}
Zhenbang Wu, Anant Dadu, Nicholas Tustison, Brian Avants, Mike Nalls, Jimeng Sun, and Faraz Faghri.
\newblock Multimodal patient representation learning with missing modalities and labels.
\newblock In \emph{The Twelfth International Conference on Learning Representations}, 2024{\natexlab{b}}.

\bibitem[Xue et~al.(2022)Xue, Gao, Ren, and Zhao]{xue2022modality}
Zihui Xue, Zhengqi Gao, Sucheng Ren, and Hang Zhao.
\newblock The modality focusing hypothesis: Towards understanding crossmodal knowledge distillation.
\newblock \emph{arXiv preprint arXiv:2206.06487}, 2022.

\bibitem[You et~al.(2020)You, Ma, Ding, Kochenderfer, and Leskovec]{you2020handling}
J. You, X. Ma, D. Ding, M.~J. Kochenderfer, and J. Leskovec.
\newblock Handling missing data with graph representation learning.
\newblock In \emph{Advances in Neural Information Processing Systems}, pages 19075--19087. NeurIPS Proceedings, 2020.

\bibitem[Zhang et~al.(2022)Zhang, Chu, Ma, Zhu, Wang, Wang, and Zhao]{zhang2022m3care}
Chaohe Zhang, Xu Chu, Liantao Ma, Yinghao Zhu, Yasha Wang, Jiangtao Wang, and Junfeng Zhao.
\newblock M3care: Learning with missing modalities in multimodal healthcare data.
\newblock In \emph{Proceedings of the 28th ACM SIGKDD Conference on Knowledge Discovery and Data Mining}, pages 2418--2428, 2022.

\bibitem[Zhang et~al.(2023)Zhang, Zhao, Liu, Wang, and Yu]{zhang2023moe}
Z. Zhang, Y. Zhao, J. Liu, S. Wang, and Q. Yu.
\newblock Moe-transformer: Robust medical image segmentation with mixture of experts.
\newblock \emph{Medical Image Analysis}, 90:\penalty0 102956, 2023.

\end{thebibliography}
}

\end{document}